\documentclass[11pt,letterpaper]{article}
\usepackage[margin=1in]{geometry}
\usepackage[T1]{fontenc}
\usepackage{lmodern}
\usepackage{amsmath}
\usepackage{amssymb}
\usepackage{amsfonts}
\usepackage{graphicx}
\usepackage{booktabs}
\usepackage{array}
\usepackage{multirow}
\usepackage{xcolor}
\usepackage{algorithm}
\usepackage{algorithmic}
\usepackage[round]{natbib}
\usepackage[hidelinks]{hyperref}

\newcommand{\appref}[1]{%
  \hyperref[#1]{Appendix~\ref*{#1}}%
}
\title{From Classifier Heads to Encoder Prototypes:\\
Personalized Federated Learning with Head-Derived Clustering}
\author{%
  Amirardalan Dehghanpour\textsuperscript{1}\thanks{Contributed equally.}\quad
  Arian Noori\textsuperscript{1}\footnotemark[1]\\[0.5em]
  Ehsan Tanghatari\textsuperscript{1}\quad
  Naimeh Omidvar\textsuperscript{2}\quad
  Hamid R. Rabiee\textsuperscript{1}\\[0.8em]
  \small \textsuperscript{1}Sharif University of Technology\\
  \small \textsuperscript{2}Tehran Institute for Advanced Studies (TeIAS)\\[0.6em]
  \small \href{mailto:ardalandehghan0@gmail.com}{\texttt{ardalandehghan0@gmail.com}}\quad
  \href{mailto:aryan1nori@gmail.com}{\texttt{aryan1nori@gmail.com}}\\
  \small \href{mailto:tanghatari.e@ipm.ir}{\texttt{tanghatari.e@ipm.ir}}\quad
  \href{mailto:omidvar@teias.institute}{\texttt{omidvar@teias.institute}}\quad
  \href{mailto:rabiee@sharif.edu}{\texttt{rabiee@sharif.edu}}%
}
\date{}
\hypersetup{
  pdftitle={From Classifier Heads to Encoder Prototypes: Personalized Federated Learning with Head-Derived Clustering},
  pdfauthor={Amirardalan Dehghanpour, Arian Noori, Ehsan Tanghatari, Naimeh Omidvar, Hamid R. Rabiee}
}
\begin{document}
\maketitle

\begin{abstract}
Personalized federated learning must decide which clients should share
representations. We introduce HDC-FL (Head-Derived Clustering for
Personalized Federated Learning), which reads that decision off a
signal shared-encoder training already produces: the client-specific
classifier heads. At periodic refresh rounds the server projects the
cached heads into a normalized low-rank descriptor space, groups the
descriptors, builds one encoder prototype per group from the cached
client encoders, and returns to each client a soft mixture of those
prototypes paired with its own unchanged head. Client-specific heads are never
averaged or distributed to other clients; the shared-encoder trajectory
is left intact; and no second optimization trajectory, additional local objective, or
post-training fine-tuning stage is required. Across 11 CIFAR-10 and
CIFAR-100 configurations covering label-subset, Dirichlet and rotation
heterogeneity with ResNet-18 and MobileNetV2 encoders, HDC-FL is
competitive with the strongest personalized and clustered baselines in
macro-averaged mean accuracy and attains the numerically highest macro-averaged worst-client and
bottom-20\% accuracy among the evaluated methods. The tail results distinguish
methods that look similar when only their means are compared. A controlled donor-swap diagnostic further finds positive donor-ranking
associations between head similarity and encoder compatibility in all
three settings tested, most clearly under covariate shift.
\end{abstract}
\section{Introduction}
\label{sec:introduction}

Federated learning (FL) trains a model from decentralized data without transferring clients' raw examples to a central server \citep{mcmahan2017,konecny2016}. In practice, however, client data are rarely independent and identically distributed. Clients may observe different label frequencies, subsets of classes, acquisition devices, or environmental conditions \citep{kairouz2021,zhao2018,hsu2019,li2020}. A model that performs well for the population average can therefore serve some clients poorly. This failure is easily obscured when models are evaluated solely by mean accuracy, which can mask severe degradation for the worst-performing clients. Personalization has also been studied as a route to more uniform performance across clients \citep{li2021ditto}. Personalized FL must consequently be judged not only by average performance, but also by how well it serves the clients that benefit least from uniform collaboration.

Personalized FL addresses this problem by allowing selected model components to vary across clients. A common decomposition shares an encoder while retaining a private classifier head for each client \citep{arivazhagan2019,collins2021,oh2022}. Although the private head adapts the decision boundary to local data, every client still relies on the same representation. This can be limiting under non-IID data: clients with different label support or acquisition conditions may require different features even when their classifiers are trained independently. Personalizing the encoder can reduce this form of negative transfer, but only if the server can determine which clients should share representation information.

Existing clustered and similarity-based methods attempt to infer such relationships, but their affinity signals may vary with individual local updates, require additional client computation or communication, or impose a hard assignment to one group. The central challenge is to obtain a stable client-relatedness signal from information already produced during training and to use it for graded representation sharing without maintaining separate optimization trajectories for different groups.

We use the geometry of client-specific classifier heads for this purpose. All client-specific heads operate on features produced along the same shared-encoder trajectory and use the same class ordering. The rows of a linear head can therefore be interpreted as client-specific class directions in a common feature space. Under label-subset skew, their arrangement reflects which class distinctions are supported by a client's data. Under covariate shift, such as image rotation, the label frequencies may remain similar while the class directions move because clients use the shared features differently. Head geometry thus provides a model-side description of the local prediction problem without requiring an additional data summary or local objective.

We introduce Head-Derived Clustering for Personalized Federated Learning (HDC-FL). At periodic refresh rounds, the server projects cached classifier heads into a normalized low-rank space and groups clients using the resulting descriptors. It then constructs prototype encoders from cached client encoders and gives each client a soft mixture of those prototypes. The personalized encoder is paired with the client's unchanged private head and is used for evaluation or deployment; it is not fed back into later federated updates. HDC-FL therefore personalizes representations while retaining one shared-encoder training trajectory. It neither averages client-specific heads nor requires a second local training stage or independently optimized cluster models.

We evaluate HDC-FL on CIFAR-10 and CIFAR-100 across 11 configurations covering label-subset, Dirichlet, and rotation heterogeneity with ResNet-18 and MobileNetV2 encoders. HDC-FL remains competitive with the strongest evaluated personalized and clustered baselines in macro-averaged mean accuracy, while providing consistently strong worst-client and bottom-$20\%$ accuracy. The tail results distinguish methods that appear similar when only their means are compared.

Our contributions are threefold:
\begin{itemize}
    \item We identify classifier-head geometry as a client-relatedness signal for representation personalization and explain how it can reflect both label and covariate heterogeneity.
    \item We develop HDC-FL, which converts low-rank head descriptors into soft mixtures of encoder prototypes without averaging client-specific heads, changing the local objective, or maintaining multiple training trajectories.
    \item We evaluate average and poorly served clients separately using mean, worst-client, and bottom-$20\%$ accuracy across 11 settings. This evaluation reveals failures hidden by mean accuracy and shows that HDC-FL combines competitive average performance with consistently strong client-level tails.
\end{itemize}
\section{Related Work}
\label{sec:related-work}

Personalized federated learning relaxes the requirement that every client use the same global model. HDC-FL connects two branches of this literature: model decomposition and similarity-based collaboration.

\paragraph{Shared representations and client-specific components.}
FedPer, FedRep, and FedBABU decompose the model into a shared feature extractor and client-specific prediction layers \citep{arivazhagan2019,collins2021,oh2022}. FedBN instead keeps batch normalization local to address feature shift \citep{li2021fedbn}, a complementary way to handle covariate heterogeneity. Private predictors adapt decision boundaries to local data, but all clients still use one shared representation. HDC-FL retains this decomposition and uses private-head geometry to determine which clients should share encoder information.

\paragraph{From hard clustering to soft association.}
Clustered FL constructs specialized models for groups of related clients \citep{sattler2021}. IFCA, PACFL, and EMD-CFL infer groups from candidate-model fit, local-data subspaces, and embedding distributions, respectively \citep{ghosh2020,vahidian2023,zhang2025}. A hard assignment is restrictive when a client can benefit from more than one group. FedSoft, FedCE, and FedAMP instead allow graded or multi-source collaboration \citep{ruan2022,cai2023,huang2021}, and pFedGraph infers graded pairwise collaboration strengths from model similarity and client data size \citep{ye2023pfedgraph}. Some such methods maintain multiple cluster models or use time-varying model evidence. HDC-FL instead mixes encoder prototypes drawn from one shared optimization trajectory.

\paragraph{Classifier-geometry-based collaboration.}
PFedCS is the closest evaluated method because it also uses classifier-parameter geometry to identify related clients \citep{wu2025}. The methods differ in what they personalize. PFedCS combines classifier parameters from similar clients, and FedPAC combines feature alignment with classifier collaboration \citep{xu2023fedpac}; both personalize the decision layer. HDC-FL instead keeps heads private, never aggregating or distributing them across clients, and uses their geometry only to mix encoder prototypes, so it personalizes the representation. In HDC-FL an encoder prototype is an averaged vector of encoder parameters; FedProto instead communicates class-level feature prototypes \citep{tan2022fedproto}.


\section{HDC-FL}
\label{sec:method}

\begin{figure*}[t]
\centering
\IfFileExists{figures/overview.jpeg}
  {\includegraphics[width=0.7\textwidth]{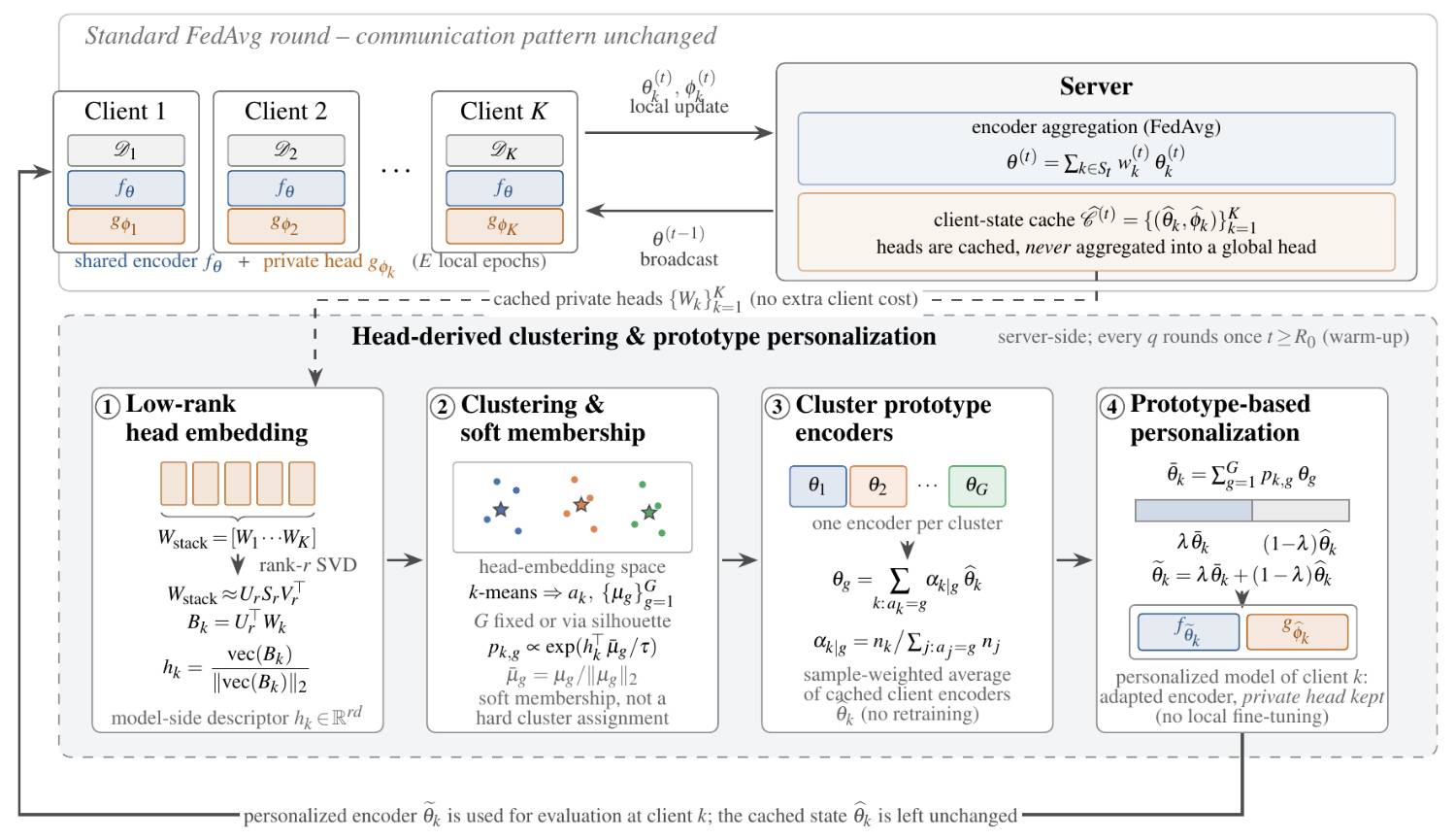}}
  {\fbox{\parbox[c][2.3cm][c]{0.7\textwidth}{\centering
   \emph{Placeholder: figures/overview.pdf (HDC-FL overview).}}}}
\caption{\textbf{HDC-FL overview.} At each refresh the server converts
cached client-specific heads into low-rank client descriptors, groups them,
constructs group-level encoder prototypes from cached client encoders,
and returns a soft prototype mixture to each client. The mixture is
paired with the client's unchanged private head, and the shared
training trajectory continues unmodified.}
\label{fig:overview}
\end{figure*}

HDC-FL adds a server-side personalization step to federated training in which the encoder is aggregated across clients while each client retains a client-specific classifier head. Participating clients return their locally updated encoder and head, and the server stores the most recent encoder–head pair available for each client. At a refresh round, the server converts the cached heads into client descriptors, groups the descriptors, constructs one encoder prototype per group from the corresponding cached encoders, and returns to each client a soft mixture of those prototypes to be used with its own head (Figure~\ref{fig:overview}). The personalization step does not modify the local objective or introduce a separate optimization trajectory.

\subsection{Setting and Cached States}
\label{sec:setting}

Client $k\in\{1,\dots,K\}$ holds
$D_k=\{(x_{k,i},y_{k,i})\}_{i=1}^{n_k}$ with $N=\sum_k n_k$. The model
pairs an encoder $f_\theta:\mathcal{X}\to\mathbb{R}^d$, shared through
aggregation, with a client-specific classifier
$g_{\phi_k}:\mathbb{R}^d\to\mathbb{R}^C$. Training minimizes
\begin{equation}
\min_{\theta,\{\phi_k\}}\ \sum_{k=1}^{K}\frac{n_k}{N}\,
\mathbb{E}_{(x,y)\sim D_k}\big[\ell(g_{\phi_k}(f_\theta(x)),y)\big].
\label{eq:objective}
\end{equation}
At round $t$ the server selects a subset $S_t$, broadcasts
$\theta^{(t-1)}$, and the selected clients train locally for $E$
epochs. Only the encoders are aggregated:
\begin{equation}
\theta^{(t)}=\sum_{k\in S_t}w_k^{(t)}\theta_k^{(t)},
\qquad w_k^{(t)}=\frac{n_k}{\sum_{j\in S_t}n_j}.
\label{eq:fedavg}
\end{equation}
In addition, the server keeps for every observed client the encoder and
the head returned by its most recent local update,
\begin{equation}
\big(\hat{\theta}_k^{(t)},\,\hat{\phi}_k^{(t)}\big),
\label{eq:cache}
\end{equation}
where we drop the round index when it is not needed. Both entries are
always refreshed together, so a cached head is the decision rule that
was trained on top of the cached encoder next to it. These pairs are
the only input to the personalization step. Under partial
participation they also keep clients that were not selected in the
current round represented, since their last available state remains in
the cache.

\subsection{Head-Derived Descriptors}
\label{sec:descriptor}

\paragraph{What head geometry encodes.}
Within a communication round, participating clients start from the same
broadcast encoder and use the same class ordering. Because local updates
in our setting are short, the resulting client encoders remain close to
the shared parameter trajectory. The coordinates of their associated
linear heads are therefore approximately aligned, allowing the rows of
$W_k \in \mathbb{R}^{C\times d}$ to be compared as client-specific class
directions in approximately common feature coordinates.

Under label-subset skew, the learned head geometry can reflect which
class distinctions are supported by a client's data; Appendix~\ref{app:donor} shows
that head similarity is strongly associated with label-support overlap
in the label-skew settings. Under covariate shift, label frequencies can
remain similar across clients while the class directions change because
clients adapt the shared representation differently. We therefore use
head geometry as a model-side description of the local prediction
problem without introducing an additional data summary or local
objective.

\paragraph{Low-rank descriptor.}
Rather than comparing the full $C\times d$ heads directly, HDC-FL
constructs a lower-dimensional representation in a common class-space
basis. Specifically, the server identifies the leading class-space
directions across the cached heads and expresses each head in that
shared basis:
\begin{equation}
\begin{aligned}
W_{\mathrm{stack}} &= [W_1\;W_2\;\cdots\;W_K]
\approx U_r\Sigma_r V_r^{\top},\\[2pt]
B_k &= U_r^{\top}W_k,\qquad
h_k = \frac{\mathrm{vec}(B_k)}{\|\mathrm{vec}(B_k)\|_2+\epsilon},
\end{aligned}
\label{eq:descriptor}
\end{equation}
where $U_r\in\mathbb{R}^{C\times r}$ holds the leading left singular
vectors of the stacked head matrix
$W_{\mathrm{stack}}\in\mathbb{R}^{C\times Kd}$ and $\epsilon>0$
prevents division by zero.

\paragraph{Descriptor compression.}
Because the heads are stacked along the feature dimension,
multiplication by $U_r^\top$ projects the class dimension of every
client's head into the same rank-$r$ basis, reducing the descriptor from
$Cd$ to $rd$ values. Appendix~\ref{app:sens} compares this representation with the
raw heads: the projected descriptors yield higher silhouette scores and
greater agreement between consecutive cluster assignments, while mean
predictive accuracy changes little. We therefore use the projection
primarily to obtain a compact and empirically more stable grouping
space.

\subsection{Clustering and Soft Membership}
\label{sec:cluster}

The server clusters the descriptors $\{h_k\}_{k=1}^{K}$ with
$k$-means, obtaining hard assignments $a_k\in\{1,\dots,G\}$ and
centroids $\{\mu_g\}_{g=1}^{G}$. The number of groups is not fixed in
advance: at every refresh $G$ is selected from the descriptor geometry
itself, so the grouping can follow the client structure as it emerges
during training instead of encoding a prior guess about how many
groups exist (Appendix~\ref{app:impl}). Writing
$\tilde{\mu}_g=\mu_g/(\|\mu_g\|_2+\epsilon)$, each client is then given
a graded membership in the groups,
\begin{equation}
p_{k,g}=\frac{\exp(h_k^{\top}\tilde{\mu}_g/\tau)}
{\sum_{j=1}^{G}\exp(h_k^{\top}\tilde{\mu}_j/\tau)} .
\label{eq:soft}
\end{equation}
The hard assignments are used \emph{only} to define distinct
prototypes; personalization itself stays soft, so a client whose
descriptor lies between two groups can draw on both. The temperature
$\tau>0$ sets how selectively a client uses the prototypes: small
$\tau$ approaches the most similar prototype, larger $\tau$ spreads
weight across groups.

\subsection{Encoder Prototypes and Personalization}
\label{sec:proto}

The grouping obtained in head space is now applied in encoder-parameter
space. Group $g$ consists of the clients assigned to it, that is all
$k$ with $a_k=g$. Its prototype $\theta_g$ is the average of the cached
encoders of exactly those clients, weighted by local dataset size so
that clients holding more data contribute more:
\begin{equation}
\theta_g=\sum_{k:\,a_k=g}\alpha_{k|g}\,\hat{\theta}_k,
\qquad
\alpha_{k|g}=\frac{n_k}{\sum_{j:\,a_j=g}n_j}.
\label{eq:proto}
\end{equation}
A prototype is therefore a single encoder that summarizes how one group
of clients has adapted the shared representation to its own data.
Client $k$ then combines the prototypes with its membership weights and
may interpolate the result with its own cached encoder:
\begin{equation}
\bar{\theta}_k=\sum_{g=1}^{G}p_{k,g}\,\theta_g,
\qquad
\tilde{\theta}_k=\lambda\,\bar{\theta}_k+(1-\lambda)\,\hat{\theta}_k,
\label{eq:mix}
\end{equation}
with $\lambda\in[0,1]$. We use $\lambda=1$ in all main experiments, so
clients receive the full soft prototype mixture; a sensitivity
diagnostic is reported in Appendix~\ref{app:sens}.

The personalized model $(f_{\tilde{\theta}_k},g_{\hat{\phi}_k})$ is
what client $k$ evaluates and deploys. It is constructed from the
cached states and is not written back into them: the encoders
aggregated in Eq.~\eqref{eq:fedavg} and the cached pairs used at the
next refresh are unchanged, so all clients continue to contribute to
one shared trajectory even though their deployed representations
differ.

\paragraph{Why cached encoders can be averaged.} The encoders in
Eq.~\eqref{eq:proto} are not independently trained networks: each local
update starts from the same broadcast encoder $\theta^{(t-1)}$ and uses
only one local epoch. Cached encoders therefore remain near a common
parameter-space trajectory, which makes weighted averaging empirically
effective in the regime we study. This argument is specific to short
local updates and frequent aggregation; longer local training or
heterogeneous architectures may weaken it, as discussed in
Section~\ref{sec:limits}.

\subsection{Refresh Schedule and Algorithm}
\label{sec:algo}

Early in training the descriptors still change quickly, so HDC-FL does
not personalize from the first round. The server repeats the refresh
every $q$ rounds and enables personalization only once the grouping has
become stable, which it measures with the silhouette score of the
current clustering together with the agreement between the assignments
of consecutive refreshes (adjusted Rand index) and the average change
of the descriptors themselves. Once this criterion has held for $m$
consecutive refreshes, prototypes are constructed and delivered; a
maximum activation round guarantees that personalization is eventually
enabled. The exact criterion and its parameter values are given in
Appendix~\ref{app:impl}, and Algorithm~\ref{alg:hdcfl} states the
complete procedure.

\begin{algorithm}[t]
\caption{HDC-FL}
\label{alg:hdcfl}
\small
\begin{algorithmic}[1]
\REQUIRE clients $\{D_k\}_{k=1}^{K}$, rounds $T$, local epochs $E$,
refresh interval $q$, rank $r$, temperature $\tau$, mixing coefficient
$\lambda$, activation parameters
\STATE initialize $\theta^{(0)}$, heads $\{\phi_k^{(0)}\}$, cache
\FOR{$t=1,\dots,T$}
  \STATE select $S_t$ and broadcast $\theta^{(t-1)}$
  \STATE each $k\in S_t$ trains locally and returns
         $(\theta_k^{(t)},\phi_k^{(t)})$
  \STATE store $(\hat{\theta}_k^{(t)},\hat{\phi}_k^{(t)})$ for
         $k\in S_t$; aggregate by Eq.~\eqref{eq:fedavg}
  \IF{$t$ is a refresh round}
    \STATE build descriptors $h_k$ from cached heads,
           Eq.~\eqref{eq:descriptor}
    \STATE select $G$ and cluster $\{h_k\}$; update the stability
           statistics
    \IF{the activation criterion is met}
      \STATE form prototypes $\theta_g$ by Eq.~\eqref{eq:proto}
      \STATE compute memberships $p_{k,g}$ by Eq.~\eqref{eq:soft}
      \STATE send each client its personalized encoder
             $\tilde{\theta}_k$, Eq.~\eqref{eq:mix}
    \ENDIF
  \ENDIF
\ENDFOR
\RETURN shared encoder $\theta^{(T)}$ and personalized models
$\{(f_{\tilde{\theta}_k},g_{\hat{\phi}_k})\}_{k=1}^{K}$
\end{algorithmic}
\end{algorithm}

\paragraph{Cost.} HDC-FL does not change the client-side objective, optimizer, or local epoch budget and requires no additional local optimization stage. Relative to shared-encoder training without head-based personalization, participating clients additionally make their client-specific heads available to the server, which stores the most recent encoder–head states required for personalization. At a refresh, the server computes a truncated SVD of the stacked heads, clusters the resulting \(K\) descriptors, constructs \(G\) weighted encoder prototypes, and computes the client-specific prototype mixtures. Personalized encoders must additionally be delivered to clients for evaluation or deployment. Thus, HDC-FL adds server-side computation, storage, and communication, but does not require independently optimized group models or additional client-side training.



\section{Experiments}
\label{sec:exp}

\subsection{Setup}
\label{sec:setup}

\paragraph{Data and heterogeneity.} We evaluate on CIFAR-10 and
CIFAR-100 \citep{krizhevsky2009} with $K=20$ clients and private
client-level train and test distributions. We consider three forms of
heterogeneity: label-subset skew (LS2/LS3), where clients observe only
20\% or 30\% of the label space; Dirichlet label skew with
$\alpha\in\{0.07,0.3\}$, which produces continuously varying label
proportions \citep{hsu2019}; and rotation shift, where clients receive one of four
fixed image rotations while label proportions follow $\mathrm{Dir}(1)$.
These settings respectively test discrete label-support differences,
continuous label skew, and client-level feature shift. Full partition
construction is given in Appendix~\ref{app:impl}.

\paragraph{Models and baselines.} We evaluate ResNet-18 and MobileNetV2
encoders \citep{he2016,sandler2018} with linear client-specific heads,
initialized from the same pretrained encoder weights within each
setting. We compare against FedAvgSharedEncoder (FedAvg-SE), which uses the same
shared-encoder/private-head training protocol without personalization;
the personalized methods FedBABU \citep{oh2022} and FedALA
\citep{zhang2023}; the clustered methods IFCA \citep{ghosh2020}, PACFL
\citep{vahidian2023} and EMD-CFL \citep{zhang2025}; the soft or
multi-cluster methods FedSoft \citep{ruan2022} and FedCE
\citep{cai2023}; and PFedCS \citep{wu2025}, the closest baseline
because it also derives client relationships from classifier geometry.
Baseline implementation and hyperparameter details are provided in
Appendix~\ref{app:impl}.

\paragraph{Protocol and metrics.} Methods are trained for 100
communication rounds, except CIFAR-100 with $\alpha=0.07$, which uses
80. Within each setting and seed, methods share the data partition,
initialization, participation schedule and local-epoch budget. The same HDC-FL configuration is used across all main benchmark
configurations; full optimization, clustering, and activation parameters
are given in Appendix~\ref{app:impl}. Ablations and sensitivity diagnostics vary the
indicated components explicitly. Results are averaged over seeds 42, 123 and
2026 and reported as mean $\pm$ sample standard deviation. We report
mean client accuracy, worst-client accuracy, bottom-20\% accuracy (the
mean of the four lowest clients) and the best-to-worst fairness gap;
aggregate values macro-average setting-level results, and ranks are
computed within a setting and then averaged. Main experiments use full
participation.

\subsection{Mean Client Accuracy}
\label{sec:mean}

Table~\ref{tab:main} reports mean client accuracy for the eight
ResNet-18 configurations; MobileNetV2 results are provided in
Appendix~\ref{app:mobilenet}. HDC-FL and FedBABU obtain the best
average rank on ResNet-18 (2.12). Across all 11 configurations, HDC-FL,
PFedCS and FedBABU achieve macro-averaged mean accuracies of 88.08\%,
87.86\% and 87.58\% respectively. Because these differences are small
relative to the per-setting seed variation, we regard the three methods
as comparable in mean accuracy rather than interpreting the aggregate
values as a clear ordering.

The most informative differences occur within individual heterogeneity
families. On CIFAR-100 label-subset skew, HDC-FL and PFedCS obtain
numerically higher mean accuracies than the remaining evaluated
baselines, reaching 88.43\% and 88.32\% on LS2 and 85.26\% and
85.45\% on LS3, respectively. HDC-FL also reaches 91.55\% under
CIFAR-10 rotation, while FedBABU has higher observed mean accuracy in
the more difficult CIFAR-100 Dirichlet and rotation settings.

\begin{table*}[t]
\caption{Mean client test accuracy (\%) on ResNet-18, mean $\pm$ sample
standard deviation over three seeds. Average rank is computed within
each setting and then averaged over the eight settings. Bold and
underline mark the best and second-best entry per setting.}
\label{tab:main}
\vskip 0.1in
\centering
\resizebox{\textwidth}{!}{%
\begin{tabular}{@{}lccccccccc@{}}
\toprule
Method & C10 LS2 & C10 LS3 & C10 Rot & C100 $\alpha{=}0.07$ & C100 $\alpha{=}0.3$ & C100 LS2 & C100 LS3 & C100 Rot & Avg. Rank $\downarrow$ \\
\midrule
FedAvg-SE & \textbf{97.17} $\pm$ 0.62 & 95.51 $\pm$ 0.71 & 86.42 $\pm$ 0.80 & 81.15 $\pm$ 1.06 & 72.09 $\pm$ 0.34 & 81.23 $\pm$ 0.63 & 76.88 $\pm$ 0.17 & 57.42 $\pm$ 0.68 & 7.25 \\
FedBABU & 96.99 $\pm$ 0.63 & 97.18 $\pm$ 1.35 & 90.05 $\pm$ 0.43 & \textbf{87.73} $\pm$ 1.51 & \textbf{82.19} $\pm$ 0.16 & 84.81 $\pm$ 4.02 & 82.33 $\pm$ 3.92 & \textbf{68.54} $\pm$ 1.91 & \textbf{2.12} \\
FedALA & 96.51 $\pm$ 0.65 & 95.50 $\pm$ 0.85 & 86.73 $\pm$ 0.03 & 81.45 $\pm$ 1.39 & 74.67 $\pm$ 0.03 & 80.96 $\pm$ 0.14 & 77.75 $\pm$ 0.40 & 63.86 $\pm$ 0.21 & 6.88 \\
FedCE & 96.89 $\pm$ 0.02 & 95.92 $\pm$ 0.12 & 86.52 $\pm$ 0.28 & 82.22 $\pm$ 0.41 & 77.19 $\pm$ 0.06 & 82.29 $\pm$ 0.76 & 78.98 $\pm$ 0.45 & 64.37 $\pm$ 0.26 & 4.50 \\
FedSoft & 96.40 $\pm$ 0.09 & 95.44 $\pm$ 1.39 & 84.41 $\pm$ 0.48 & 81.90 $\pm$ 0.94 & 75.41 $\pm$ 0.22 & 81.63 $\pm$ 0.25 & 79.23 $\pm$ 0.04 & 61.06 $\pm$ 0.59 & 6.88 \\
IFCA & 96.36 $\pm$ 1.05 & 94.38 $\pm$ 0.42 & 87.18 $\pm$ 0.13 & 80.12 $\pm$ 0.28 & 75.93 $\pm$ 0.55 & 80.61 $\pm$ 1.00 & 78.40 $\pm$ 0.23 & 63.69 $\pm$ 0.46 & 7.62 \\
PACFL & 96.69 $\pm$ 0.35 & 95.35 $\pm$ 0.13 & 86.43 $\pm$ 0.29 & 82.87 $\pm$ 0.62 & 76.35 $\pm$ 0.27 & 81.77 $\pm$ 0.11 & 78.18 $\pm$ 0.43 & \underline{64.57} $\pm$ 0.65 & 5.50 \\
EMD-CFL & 96.70 $\pm$ 0.14 & 94.88 $\pm$ 0.30 & 82.16 $\pm$ 1.86 & 80.23 $\pm$ 0.28 & 70.84 $\pm$ 1.70 & 79.85 $\pm$ 1.37 & 77.04 $\pm$ 0.74 & 60.76 $\pm$ 4.74 & 8.88 \\
PFedCS & 96.34 $\pm$ 0.51 & \underline{97.62} $\pm$ 0.64 & \underline{90.43} $\pm$ 0.15 & \underline{87.12} $\pm$ 1.09 & \underline{80.19} $\pm$ 0.05 & \underline{88.32} $\pm$ 0.27 & \textbf{85.45} $\pm$ 0.15 & 64.05 $\pm$ 0.82 & 3.25 \\
\textbf{HDC-FL} & \underline{96.94} $\pm$ 0.04 & \textbf{97.88} $\pm$ 0.60 & \textbf{91.55} $\pm$ 0.04 & 86.77 $\pm$ 0.54 & 79.84 $\pm$ 0.18 & \textbf{88.43} $\pm$ 0.29 & \underline{85.26} $\pm$ 0.14 & 64.38 $\pm$ 0.14 & \textbf{2.12} \\
\bottomrule
\end{tabular}}
\end{table*}

\subsection{Client-Level Tail Performance}
\label{sec:tail}

Mean accuracy can conceal clients that are served poorly by a
personalized method. Table~\ref{tab:tail} therefore macro-averages
worst-client accuracy, bottom-20\% accuracy, fairness gap and
bottom-20\% rank across all 11 configurations; setting-level results
are reported in Appendix~\ref{app:tailfull}.

\begin{table}[tb]
\caption{Client-level results macro-averaged over all 11
configurations.}
\label{tab:tail}
\vskip 0.1in
\centering
\footnotesize
\setlength{\tabcolsep}{3pt}
\begin{tabular}{@{}lccccc@{}}
\toprule
Method & Mean & Worst & Bot.20\% & Gap & Tail rank \\
\midrule
FedAvg-SE & 83.35 & 76.95 & 78.95 & 11.15 & 6.00 \\
FedBABU & 87.58 & 55.32 & 73.78 & 38.37 & 8.18 \\
FedALA & 84.19 & 77.68 & 79.75 & 11.07 & 5.73 \\
FedCE & 85.03 & 78.73 & 80.69 & 10.94 & 3.73 \\
FedSoft & 84.15 & 76.52 & 79.31 & 12.38 & 6.00 \\
IFCA & 83.64 & 71.25 & 77.58 & 17.90 & 7.27 \\
PACFL & 84.35 & 75.38 & 79.39 & 14.26 & 5.00 \\
EMD-CFL & 82.43 & 71.72 & 75.45 & 16.65 & 8.36 \\
PFedCS & \underline{87.86} & \underline{81.69} & \underline{84.20} & \underline{9.74} & \underline{2.64} \\
\textbf{HDC-FL} & \textbf{88.08} & \textbf{82.05} & \textbf{84.44} & \textbf{9.72} & \textbf{2.09} \\
\bottomrule
\end{tabular}
\end{table}

HDC-FL attains the highest aggregate worst-client accuracy (82.05\%)
and bottom-20\% accuracy (84.44\%), with the best tail rank (2.09).
However, its margins over PFedCS are small---0.36 and 0.24 percentage
points on the two tail metrics---and their fairness gaps are nearly
identical (9.72 versus 9.74). We therefore view the two
classifier-geometry methods as similarly strong on client-level tails;
both remain clearly separated from the remaining baselines in aggregate
tail rank and worst-client accuracy.

FedBABU illustrates why the tail metrics add information beyond the
mean: despite a competitive aggregate mean of 87.58\%, its worst-client
accuracy falls to 55.32\% and its fairness gap rises to 38.37 points.
Hard-clustering methods also show weaker tails than their means alone
suggest: IFCA and EMD-CFL both fall below the non-personalized
FedAvg-SE baseline in aggregate worst-client accuracy. These
results suggest that personalization mechanisms can perform well for
the average client while harming a subset of clients.

\subsection{Ablations}
\label{sec:ablation}

We isolate two central design choices: whether head-derived grouping
improves over population-level prototype averaging, and whether soft
prototype membership improves over hard assignment. Additional
descriptor and hyperparameter diagnostics are reported in
Appendix~\ref{app:sens}.

\paragraph{Grouping signal.} Table~\ref{tab:abl-group} isolates the two
steps that build the personalized encoder, on CIFAR-10 rotation without
pretrained initialization. Replacing the client's own cached encoder by
a single population-level prototype already raises mean accuracy from
77.57\% to 80.30\%, so part of the benefit is simply that a prototype
averages several cached encoders. Splitting that prototype into four
head-derived groups adds a further 0.74 points in mean accuracy and
also improves both tail metrics. Thus, after accounting for the benefit
of prototype averaging itself, introducing multiple head-derived
prototypes provides an additional improvement on all three metrics.
Additional diagnostics in Appendix~\ref{app:sens} show that the
low-rank projection substantially improves cluster separation and
temporal stability relative to raw heads without materially changing
predictive accuracy.

\begin{table}[tb]
\caption{Controlled grouping ablation on CIFAR-10 rotation without pretrained
initialization. Here $G=4$ is fixed to match the four known rotation
conditions and isolate the effect of head-derived grouping. Values are mean
$\pm$ sample standard deviation over three seeds.}
\label{tab:abl-group}
\vskip 0.1in
\centering

\resizebox{\columnwidth}{!}{%
\begin{tabular}{@{}lccc@{}}
\toprule
Variant & Mean & Worst & Bot.20\% \\
\midrule
No prototype
& 77.57 $\pm$ 0.40
& 69.85 $\pm$ 3.97
& 72.20 $\pm$ 1.46 \\

One prototype
& 80.30 $\pm$ 0.34
& 72.05 $\pm$ 2.90
& 74.93 $\pm$ 1.16 \\

Head-derived, $G{=}4$
& \textbf{81.04} $\pm$ 0.73
& \textbf{72.57} $\pm$ 1.67
& \textbf{75.47} $\pm$ 1.10 \\
\bottomrule
\end{tabular}%
}
\end{table}

\paragraph{Membership.} Replacing the soft mixture by a hard
assignment to the most similar prototype costs 0.85 points of mean
accuracy and 1.08 and 0.95 points on the two tail metrics on CIFAR-100
with $\alpha=0.3$, the setting in which client distributions form a
continuum rather than discrete groups (Appendix~\ref{app:sens}).

\subsection{Does Head Similarity Predict Encoder Compatibility?}
\label{sec:donor}

We also test the central assumption of the method directly: that
clients with similar classifier heads can use related encoders. At the final round of the MobileNetV2 runs we
take every ordered pair of clients $(t,d)$, keep target $t$'s own
classifier head and test data fixed, replace only its encoder with
donor $d$'s cached encoder, and re-evaluate. The swap is zero-shot, with no
retraining or fine-tuning. Each pair yields a loss benefit and an accuracy gain relative to
the target's own encoder, and we ask how well head similarity ranks the
19 available donors by these quantities. Three settings and three seeds
give 3{,}420 directed pairs; metrics are computed per target and then
averaged (full protocol in Appendix~\ref{app:donor}).

Table~\ref{tab:donor} reports the outcome. Head similarity has positive mean
Spearman correlation with loss benefit and AUROC above $0.5$ in all
three settings, with the largest observed association under rotation
shift ($\rho=0.243$, AUROC $=0.783$) and the smallest under CIFAR-100
LS3 ($\rho=0.117$, AUROC $=0.565$). The practical effect is measured by
donor selection: choosing the nearest-head donor rather than a random
one is worth $2.50$ accuracy points under rotation, $1.57$ points on
CIFAR-10 label skew and $0.41$ points on CIFAR-100 label skew, and the
closest distance quintile outperforms the farthest in every setting.
Average precision must be read against prevalence: on CIFAR-100 label
skew $71.1\%$ of donors already help, so an AP of $0.782$ is a lift of
only $0.071$.

The rotation setting provides the cleanest check, because the client
groups are known independently of the learned heads. The nearest-head
donor shares the target's rotation in $75.0\%$ of cases against a
$21.1\%$ base rate, head similarity separates same-rotation from
different-rotation pairs with AUROC $0.887$, and same-rotation swaps
average $+0.83$ accuracy points against $-3.46$ points for
different-rotation swaps. Two qualifications follow from the same
experiment. First, a useful ranking is not a guarantee of useful
transfer: on CIFAR-10 label skew only $15.6\%$ of individual swaps
reduce the target's loss, so most single-donor replacements are harmful
even where the ranking is informative. This is consistent with HDC-FL
combining several cached encoders by sample-weighted averaging and soft
mixing rather than adopting one donor's encoder. Second, in the
label-skew settings head similarity is closely associated with the
overlap of client label supports (Spearman $0.89$ and $0.78$), and the
partial correlation with transfer benefit after controlling for that
overlap is near zero (Appendix~\ref{app:donor}). In these settings the
signal is therefore better described as an operational proxy for shared
label support than as an independent measure of encoder compatibility.

\begin{table}[tb]
\caption{
Donor-compatibility diagnostic on MobileNetV2, averaged over targets
and seeds. Spearman $\rho$ measures rank association between head
similarity and loss benefit, while AUROC and average precision (AP)
measure discrimination between beneficial and non-beneficial donors.
The beneficial-donor rate gives the positive-class prevalence against
which AP should be interpreted. The final two rows report the accuracy
gain from selecting the nearest-head donor rather than a random donor,
and from the closest head-distance quintile relative to the farthest.
}
\label{tab:donor}
\vskip 0.1in
\centering
\footnotesize
\setlength{\tabcolsep}{4pt}
\begin{tabular}{@{}lccc@{}}
\toprule
 & C10 LS3 & C100 LS3 & C100 Rot \\
\midrule
Spearman $\rho$ (loss)      & 0.193 & 0.117 & \textbf{0.243} \\
AUROC (loss)                & 0.665 & 0.565 & \textbf{0.783} \\
Beneficial-donor rate       & 0.156 & 0.711 & 0.245 \\
AP                          & 0.513 & 0.782 & 0.642 \\
AP lift over prevalence     & 0.357 & 0.071 & \textbf{0.397} \\
Top-1 $-$ random (pp)       & 1.57 & 0.41 & \textbf{2.50} \\
Closest $-$ farthest (pp)   & 1.68 & 0.51 & \textbf{1.97} \\
\bottomrule
\end{tabular}
\end{table}

\section{Discussion}
\label{sec:disc}

HDC-FL is not uniformly more accurate than the strongest evaluated baselines. The results instead support a narrower claim: classifier-head geometry is a useful model-side signal for deciding how to share representations. Cross-method comparisons cannot isolate this signal from other algorithmic differences, so the donor-swap diagnostic in Section~\ref{sec:donor} tests the underlying association directly. Head similarity positively ranks encoder-transfer benefit in all three tested settings and recovers the known rotation structure under covariate shift. Under label skew, Appendix~\ref{app:donor} shows that much of this association is explained by shared label support.

\subsection{Limitations}
\label{sec:limits}

Two assumptions bound the setting. The first is architectural: all
clients train the same encoder over a common label space, since
descriptors compare heads coordinate by coordinate and prototypes
average encoders parameter by parameter. The argument that those
averages are meaningful further assumes short local updates and
frequent aggregation, and longer local training or heterogeneous client
models would weaken it; relaxing both is left to future work.
The second assumption is informational: in the current implementation,
the server receives the client-specific classifier heads, which can
encode information about client-specific prediction behavior. HDC-FL
as evaluated here is therefore not suited to settings in which model
parameters themselves must remain hidden from the server. After the
descriptor construction step, clustering and soft-membership
computation operate on the descriptors rather than directly on the raw
heads, suggesting privacy-preserving computation or transmission of
descriptor statistics as possible extensions. We do not evaluate such
mechanisms here. The donor
diagnostic of Section~\ref{sec:donor} is correlational, uses one
backbone, and shows head similarity partly acting as a proxy for
label-support overlap. A third direction is
scale: our federations have 20 clients, and testing
substantially larger and naturally partitioned populations, where the
server cache grows and each client is refreshed less often, is the
natural next step.

\section{Conclusion}

We introduced HDC-FL, which uses the geometry of the private classifier
heads that shared-encoder training already produces to decide how
representations are shared, returning to each client a soft mixture of
head-derived encoder prototypes paired with its own unchanged head.
Across 11 configurations and two backbones, HDC-FL is competitive with
the strongest baselines in mean accuracy and attains the highest
macro-averaged worst-client and bottom-20\% accuracy, with small margins
over PFedCS and no additional client-side optimization. The contrast with
FedBABU---a comparable mean but a 38-point fairness gap---illustrates
why personalized federated learning should also be evaluated by
client-level tails.

\clearpage
\section*{Supplementary Materials}
\appendix

\section{EXPERIMENTAL DETAILS}
\label{app:impl}

\paragraph{Partitions.} Every configuration uses $K=20$ clients. Under
label-subset skew each client draws a random subset of the classes
containing 20\% of them (LS2) or 30\% of them (LS3)---two or three
classes on CIFAR-10 and twenty or thirty on CIFAR-100---and the
examples of each class are divided only among the clients that own that
class. Under Dirichlet skew the class proportions of each client are
drawn from $\mathrm{Dir}(\alpha)$ with $\alpha\in\{0.07,0.3\}$. Under
rotation shift each client is assigned one of $0^{\circ}$, $90^{\circ}$,
$180^{\circ}$, $270^{\circ}$, giving five clients per rotation, and its
label proportions are drawn from $\mathrm{Dir}(1)$.

\paragraph{Train and test splits.} We use the official CIFAR splits and
do not resample them. The 50{,}000 training images are partitioned
across the 20 clients by the procedure above, and the 10{,}000 test
images are then allocated class by class in proportion to the clients'
training counts for that class: a client holding a given fraction of
the training examples of class $c$ receives the same fraction of the
test examples of class $c$, and a class no client trains on is given to
no client. Each client therefore has a private test split that follows
its own training distribution, and every reported accuracy is computed
on that split.

\paragraph{Training and preprocessing.}
All methods run for 100 communication rounds, except CIFAR-100 with
$\alpha=0.07$, which runs for 80. Unless otherwise required by a
method-specific optimization stage, client updates use AdamW with one
local epoch per communication round, batch size 20, encoder and head
learning rates of $10^{-3}$, weight decay $10^{-4}$, and no
learning-rate scheduler.

Training images are processed using ImageNet-style augmentation: a
$224\times224$ random resized crop with scale $[0.2,1.0]$, random
horizontal flipping, and ImageNet normalization. Test images are resized
and center-cropped to $224\times224$ and use the same normalization.
The same preprocessing is used across methods. Method-specific
optimization stages, including FedBABU personalization and PFedCS
classifier fine-tuning, are specified below.

\paragraph{Initialization.} Encoders start from the torchvision
ImageNet-1k checkpoints \citep{russakovsky2015}: \texttt{ResNet18\_Weights.DEFAULT} for
ResNet-18 and \texttt{MobileNet\_V2\_Weights.IMAGENET1K\_V1} for
MobileNetV2. The classifier heads are single linear layers and are
always randomly initialized. Within each setting and seed, all methods use the same pretrained
encoder initialization and the same randomly initialized client heads. The diagnostics in Table~\ref{tab:abl-group} and
Appendix~\ref{app:sens} instead start from randomly initialized
encoders and are labeled as such.

\paragraph{Asset provenance and terms.}
CIFAR-10 and CIFAR-100 were obtained from their official public
distribution page, which requests citation of the accompanying
technical report but does not state a separate dataset license
(\url{https://www.cs.toronto.edu/~kriz/cifar.html}). Model definitions
and the ImageNet-1k checkpoints named above were obtained through
torchvision. The torchvision source code is distributed under the
BSD 3-Clause license; its documentation notes that pretrained weights
may also be subject to terms inherited from their training data. We use
the ImageNet-pretrained weights only for non-commercial research,
consistent with the ImageNet terms of access
(\url{https://www.image-net.org/accessagreement}). All baseline
algorithms were independently reimplemented in our common experimental
framework from their published descriptions; no third-party baseline
repository code was incorporated.

\paragraph{Hardware.} Runs were executed on a single NVIDIA Quadro RTX
6000 GPU, with one run scheduled at a time.

\paragraph{HDC-FL configuration.} Descriptor rank $r=8$, temperature
$\tau=0.1$, mixing coefficient $\lambda=1$, refresh interval $q=1$, so
that descriptors, groups and prototypes are rebuilt at every round once
personalization is active, candidate cluster range
$G\in\{2,\dots,8\}$, and
activation parameters $R_{\min}=8$, $R_{\max}=30$, $m=6$,
$\rho_{\min}=0.80$, $\delta_{\max}=0.02$, $s_{\min}=0.24$. The same configuration is used across all main benchmark settings, without per-setting retuning. Appendix~\ref{app:sens} reports sensitivity and diagnostic analyses around this fixed configuration.


\subsection{Dynamic Cluster Selection}
\label{app:dynamic-clusters}

HDC-FL does not fix the number of prototype groups in advance. At each
refresh round, the server selects the number of groups directly from the
current geometry of the normalized head-derived descriptors
$\{h_k\}_{k=1}^{K}$. Let
\begin{equation}
\mathcal{G}
=
\{G_{\min},G_{\min}+1,\ldots,G_{\max}\}
\label{eq:cluster-candidates}
\end{equation}
denote the candidate set of cluster counts. For every
$G\in\mathcal{G}$, the server applies $k$-means to the descriptors and
obtains assignments
$a_k^{(G)}\in\{1,\ldots,G\}$ and centroids
$\{\mu_g^{(G)}\}_{g=1}^{G}$.

Because the descriptors are $\ell_2$-normalized, cluster quality is
measured using cosine distance,
\begin{equation}
d(h_i,h_j)
=
1-h_i^{\top}h_j .
\label{eq:cosine-distance}
\end{equation}
For a client $k$ under a candidate clustering with $G$ groups, let
$d^{\mathrm{in}}_{k,G}$ denote its average distance to the other
descriptors in the same cluster, and let
$d^{\mathrm{out}}_{k,G}$ denote the minimum, over all other clusters, of
its average distance to the descriptors in that cluster. The silhouette
score \citep{rousseeuw1987} of the candidate clustering is
\begin{equation}
\operatorname{Sil}(G)
=
\frac{1}{K}
\sum_{k=1}^{K}
\frac{
d^{\mathrm{out}}_{k,G}
-
d^{\mathrm{in}}_{k,G}
}{
\max\{
d^{\mathrm{in}}_{k,G},
d^{\mathrm{out}}_{k,G}
\}
}.
\label{eq:silhouette}
\end{equation}
The number of prototype groups used at that refresh is then
\begin{equation}
\widehat{G}
=
\arg\max_{G\in\mathcal{G}}
\operatorname{Sil}(G).
\label{eq:dynamic-G}
\end{equation}

In all experiments we use
$G_{\min}=2$ and $G_{\max}=8$. For every candidate value, $k$-means is
run with multiple random initializations, and the candidate attaining the
largest silhouette score is selected. Thus, the number of prototypes can
change across refresh rounds as the client-descriptor geometry evolves.
The resulting hard assignments are used only to construct distinct
encoder prototypes; client personalization remains soft through the
prototype-membership weights defined in the main text.

\subsection{Stability-Based Warmup and Activation}
\label{app:warmup}

Head-derived descriptors can be unstable early in training because both
the shared encoder and the client-specific classifier heads are still
adapting. Constructing personalized encoder prototypes from this
transient geometry may therefore produce unstable client groups.
HDC-FL consequently begins with a warmup phase during which the
descriptor geometry is monitored but prototype-based personalization is
not yet activated. During this phase, training follows the shared
encoder trajectory described in the main text.

Let
$a^{(t)}=(a_1^{(t)},\ldots,a_K^{(t)})$
denote the client assignments obtained at refresh round $t$ after
dynamically selecting $\widehat{G}^{(t)}$. We measure assignment
consistency between consecutive refreshes using the adjusted Rand index
(ARI),
\begin{equation}
\rho^{(t)}
=
\operatorname{ARI}
\left(
a^{(t-1)},a^{(t)}
\right).
\label{eq:warmup-ari}
\end{equation}
Larger values indicate greater agreement between consecutive
clusterings.

We additionally measure the movement of the descriptor space itself.
Because each descriptor is $\ell_2$-normalized, the average cosine
drift between consecutive refreshes is
\begin{equation}
\Delta^{(t)}
=
\frac{1}{K}
\sum_{k=1}^{K}
\left[
1-
\left(h_k^{(t-1)}\right)^{\top}
h_k^{(t)}
\right].
\label{eq:descriptor-drift}
\end{equation}
Smaller $\Delta^{(t)}$ indicates that the descriptor geometry is
changing little, even when individual assignments lie near cluster
boundaries.

Let
\begin{equation}
s^{(t)}
=
\operatorname{Sil}^{(t)}
\left(
\widehat{G}^{(t)}
\right)
\label{eq:selected-silhouette}
\end{equation}
denote the silhouette score of the dynamically selected clustering at
refresh round $t$. Following the implementation, a refresh is declared
stable when the selected clustering has sufficient separation and
either its assignments are temporally consistent or the descriptor
geometry has changed only slightly:
\begin{equation}
\mathcal{S}^{(t)}
=
\left[
s^{(t)} \ge s_{\min}
\right]
\land
\left(
\left[
\rho^{(t)} \ge \rho_{\min}
\right]
\lor
\left[
\Delta^{(t)} \le \delta_{\max}
\right]
\right).
\label{eq:stability-criterion}
\end{equation}
Thus, sufficient cluster separation is always required. The second
condition permits stability to be established either through persistent
cluster assignments or through small descriptor drift. In particular,
small descriptor drift alone does not trigger personalization when the
current clustering is poorly separated.

To avoid activation because of a single favorable refresh,
$\mathcal{S}^{(t)}=1$ must hold for $m$ consecutive refresh checks after
a minimum warmup round $R_{\min}$. Let $t_{\mathrm{act}}$ denote the
round at which prototype personalization becomes active. In the main
experiments, where descriptors are refreshed every round ($q=1$), this
is
\begin{equation}
t_{\mathrm{act}}
=
\min
\left\{
R_{\max},
\;
\min
\left\{
t \ge R_{\min} :
\mathcal{S}^{(t-m+1)}
=
\cdots
=
\mathcal{S}^{(t)}
=
1
\right\}
\right\}.
\label{eq:activation-round}
\end{equation}
The maximum activation round $R_{\max}$ is a fallback that activates
personalization if the stability criterion has not held for $m$
consecutive refreshes by that point.

Before $t_{\mathrm{act}}$, clients are evaluated using the shared
encoder trajectory together with their client-specific heads, and no
prototype-based personalized encoder is deployed. At and after
$t_{\mathrm{act}}$, the server constructs encoder prototypes and
client-specific soft mixtures as described in
Section~\ref{sec:proto}. The
descriptors, dynamically selected group count, assignments,
memberships, and prototypes are subsequently recomputed every $q$
refresh rounds.

For all main experiments we use
\begin{equation}
R_{\min}=8,\qquad
R_{\max}=30,\qquad
m=6,\qquad
\rho_{\min}=0.80,\qquad
\delta_{\max}=0.02,\qquad
s_{\min}=0.24.
\label{eq:activation-hyperparameters}
\end{equation}
The same activation configuration is used across all main benchmark
settings.
\paragraph{Baseline implementation.}
We implement all baselines within the same experimental framework so
that methods share the client partitions, backbone initialization,
participation schedule, communication-round budget, and local training
budget within each setting and seed. Unless a method requires an
additional method-specific optimization stage, client updates follow
the common optimization protocol described above. Method-specific
mechanisms and hyperparameters are based on the corresponding papers
or public implementations where applicable, but are adapted to this
common training protocol rather than reproducing each method's
original experimental budget. Table~\ref{tab:baseline-configs} lists
the exact method-specific settings used in our experiments. The same
method-specific configuration is used across benchmark settings unless
stated otherwise.

\begin{table*}[t]
\centering
\caption{
Method-specific baseline configurations used in our experiments. FedAvg-SE denotes FedAvgSharedEncoder.
}
\label{tab:baseline-configs}
\small
\setlength{\tabcolsep}{5pt}
\renewcommand{\arraystretch}{1.25}
\begin{tabular}{p{0.15\textwidth} p{0.79\textwidth}}
\toprule
\textbf{Method} & \textbf{Method-specific configuration} \\
\midrule

FedAvg-SE &
The encoder is aggregated across participating clients using
sample-count weighting, while each client retains its own classifier
head. No additional personalization stage is applied. \\

FedBABU &
The classifier is kept fixed during federated training and only the
encoder/body is locally optimized and aggregated. Final personalization
uses 5 epochs of full-model fine-tuning with SGD, learning rate
$10^{-2}$, momentum $0$, and weight decay $0$. \\

FedALA &
ALA depth $p=2$, local-data fraction $s=80\%$, and weight-learning rate
$\eta=1$. Initial ALA weight optimization is capped at 50 epochs with
patience 3 and tolerance $10^{-4}$; subsequent rounds use one ALA
weight-update epoch. \\

IFCA &
Uses $S=4$ cluster models. Client-to-cluster assignment is based on
model loss evaluated on at most 512 local samples. \\

PACFL &
Client subspace dimension $p=3$, hierarchical-clustering threshold
$1.5$ radians, average linkage, and one-shot client clustering. \\

FedSoft &
Uses $S=3$ cluster centers, proximal coefficient $\lambda=0.1$,
importance-estimation interval $\tau=2$, smoothing constant
$\sigma=10^{-4}$, and client-selection size
$K_{\mathrm{sel}}=20$. \\

FedCE &
Uses $M=4$ cluster models, initial coefficient $\delta_0=0.5$ with
$\delta(t)=\delta_0/(1+t)$, identity-set size 2, softmax scale 1, and
proximal coefficient $0.01$. \\

EMD-CFL &
Uses EMD tolerance $0.025$, projection ratio $0.1$, and one-shot
clustering at round 0. \\

PFedCS &
Uses stage boundary $\beta=T/2$, classifier fine-tuning for
$\rho=1$ local epoch, collaboration mixture coefficient $0.5$,
knowledge-distillation temperature 1, and
$\eta_v=\eta_w=10^{-3}$ under the common optimization protocol. \\

\bottomrule
\end{tabular}
\end{table*}

\paragraph{Metrics.} For final client accuracies $a_1,\dots,a_K$, the
mean client accuracy is $\frac{1}{K}\sum_k a_k$, the worst-client
accuracy is $\min_k a_k$, the bottom-20\% accuracy is the mean of the
$\max(1,\lceil 0.2K\rceil)$ lowest values, and the fairness gap is
$\max_k a_k-\min_k a_k$. Within a setting, methods are ranked by the
metric of interest with average ranks for ties; aggregate ranks
macro-average the setting-level ranks, giving every configuration equal
weight.

\section{MOBILENETV2 MEAN ACCURACY}
\label{app:mobilenet}

Table~\ref{tab:mobilenet} completes the 11-configuration mean-accuracy
evaluation with the three MobileNetV2 settings omitted from
Table~\ref{tab:main}.

\begin{table}[!htbp]
\caption{MobileNetV2 mean client test accuracy (\%), mean $\pm$ sample
standard deviation over seeds 42, 123 and 2026. Bold and underline mark
the best and second-best entry per setting.}
\label{tab:mobilenet}
\vskip 0.1in
\centering
\setlength{\tabcolsep}{2.5pt}
\renewcommand{\arraystretch}{1.05}
\resizebox{\linewidth}{!}{%
\begin{tabular}{@{}lcccc@{}}
\toprule
Method & CIFAR-100 LS3 & CIFAR-10 $\alpha{=}0.07$ & CIFAR-10 LS3 & Avg. Rank $\downarrow$ \\
\midrule
FedAvg-SE & 77.43 $\pm$ 0.37 & 95.85 $\pm$ 0.28 & 95.25 $\pm$ 0.74 & 6.67 \\
FedBABU & 81.27 $\pm$ 1.89 & \textbf{96.26} $\pm$ 1.19 & 96.05 $\pm$ 1.42 & \textbf{2.33} \\
FedALA & 77.77 $\pm$ 0.27 & 95.99 $\pm$ 0.10 & 94.87 $\pm$ 1.17 & 5.67 \\
FedCE & 78.87 $\pm$ 0.43 & \underline{96.12} $\pm$ 0.38 & 95.96 $\pm$ 0.51 & 4.00 \\
FedSoft & 79.57 $\pm$ 0.48 & 95.07 $\pm$ 1.13 & 95.52 $\pm$ 0.39 & 5.33 \\
IFCA & 77.69 $\pm$ 0.49 & 90.90 $\pm$ 1.54 & 94.83 $\pm$ 0.65 & 8.67 \\
PACFL & 78.87 $\pm$ 0.48 & 92.74 $\pm$ 2.24 & 94.05 $\pm$ 3.32 & 7.67 \\
EMD-CFL & 77.50 $\pm$ 1.94 & 92.44 $\pm$ 0.55 & 94.36 $\pm$ 0.80 & 9.00 \\
PFedCS & \textbf{84.93} $\pm$ 0.06 & 95.20 $\pm$ 0.87 & \underline{96.81} $\pm$ 0.78 & 3.00 \\
\textbf{HDC-FL} & \underline{84.66} $\pm$ 0.42 & 95.70 $\pm$ 0.72 & \textbf{97.43} $\pm$ 0.58 & \underline{2.67} \\
\bottomrule
\end{tabular}}
\end{table}

\section{SETTING-LEVEL TAIL RESULTS}
\label{app:tailfull}

Tables~\ref{tab:worst}, \ref{tab:bottom} and~\ref{tab:gap} give the
per-configuration values behind the macro-averages of
Table~\ref{tab:tail}: worst-client accuracy, bottom-20\% accuracy and
fairness gap. Columns prefixed with ``M-'' are MobileNetV2
configurations; all others use ResNet-18. ``Rot'' denotes rotation
shift with $\mathrm{Dir}(1)$ label proportions.

Two patterns are easier to see here than in the aggregate. First, the
seed-to-seed variability of the tail differs enormously between
methods: FedBABU's worst-client accuracy has standard deviations above
20 points in four configurations, while the two classifier-geometry
methods stay below 4 points in all but one. Second, the fairness gaps
of HDC-FL and PFedCS are close in every configuration, not only on
average, which is why we treat their aggregate difference of 0.02
points as a tie rather than as an ordering.

\begin{table}[!htbp]
\caption{Final worst-client test accuracy (\%), mean $\pm$ sample
standard deviation over three seeds.}
\label{tab:worst}
\vskip 0.1in
\centering
\setlength{\tabcolsep}{2.5pt}
\renewcommand{\arraystretch}{1.05}
\resizebox{\linewidth}{!}{%
\begin{tabular}{@{}lccccccccccc@{}}
\toprule
Method & C10 LS2 & C10 LS3 & C10 Rot & C100 $\alpha{=}.07$ & C100 $\alpha{=}.3$ & C100 LS2 & C100 LS3 & C100 Rot & M-C100 LS3 & M-C10 $\alpha{=}.07$ & M-C10 LS3 \\
\midrule
FedAvg-SE & 88.73 $\pm$ 5.75 & 90.23 $\pm$ 1.03 & 79.53 $\pm$ 4.86 & 74.29 $\pm$ 0.82 & 67.10 $\pm$ 0.69 & 75.70 $\pm$ 0.69 & 68.06 $\pm$ 1.50 & 52.29 $\pm$ 1.28 & 72.45 $\pm$ 0.60 & 87.92 $\pm$ 2.67 & 90.17 $\pm$ 0.18 \\
FedBABU & 77.64 $\pm$ 10.88 & 82.66 $\pm$ 13.87 & 75.68 $\pm$ 3.07 & 55.35 $\pm$ 26.75 & 34.07 $\pm$ 6.89 & 51.94 $\pm$ 31.17 & 35.67 $\pm$ 5.29 & 20.04 $\pm$ 2.68 & 27.40 $\pm$ 6.79 & 76.28 $\pm$ 24.78 & 71.84 $\pm$ 18.52 \\
FedALA & 86.89 $\pm$ 7.23 & 90.20 $\pm$ 2.12 & 80.41 $\pm$ 4.74 & 72.97 $\pm$ 1.82 & 68.51 $\pm$ 1.17 & 74.21 $\pm$ 2.59 & 72.82 $\pm$ 0.84 & 57.50 $\pm$ 0.26 & 72.57 $\pm$ 0.79 & \textbf{88.87} $\pm$ 0.48 & 89.53 $\pm$ 2.33 \\
FedCE & 88.33 $\pm$ 5.50 & 91.50 $\pm$ 0.28 & 79.46 $\pm$ 4.10 & 75.33 $\pm$ 1.96 & 72.30 $\pm$ 1.19 & 75.92 $\pm$ 0.07 & 72.29 $\pm$ 2.25 & 57.76 $\pm$ 0.76 & 73.53 $\pm$ 1.19 & 88.28 $\pm$ 2.07 & 91.36 $\pm$ 0.79 \\
FedSoft & 87.55 $\pm$ 4.39 & 87.73 $\pm$ 3.01 & 75.76 $\pm$ 4.87 & 76.05 $\pm$ 0.55 & 66.92 $\pm$ 0.58 & 75.64 $\pm$ 0.14 & 72.14 $\pm$ 0.69 & 54.98 $\pm$ 1.36 & 73.90 $\pm$ 1.97 & 82.63 $\pm$ 6.61 & 88.40 $\pm$ 2.80 \\
IFCA & 86.15 $\pm$ 8.06 & 69.67 $\pm$ 26.74 & 81.83 $\pm$ 2.86 & 73.27 $\pm$ 0.01 & 69.68 $\pm$ 0.27 & 73.11 $\pm$ 0.49 & 68.78 $\pm$ 8.28 & 56.89 $\pm$ 3.31 & 70.65 $\pm$ 2.51 & 47.28 $\pm$ 13.44 & 86.41 $\pm$ 4.12 \\
PACFL & 88.11 $\pm$ 3.30 & 88.56 $\pm$ 0.38 & 78.58 $\pm$ 0.39 & 76.60 $\pm$ 0.91 & 70.32 $\pm$ 1.05 & 75.94 $\pm$ 0.66 & 72.72 $\pm$ 0.80 & 59.16 $\pm$ 0.74 & 72.70 $\pm$ 1.93 & 55.03 $\pm$ 27.38 & 91.49 $\pm$ 2.64 \\
EMD-CFL & \textbf{90.67} $\pm$ 0.63 & 88.90 $\pm$ 1.81 & 73.43 $\pm$ 3.05 & 69.82 $\pm$ 4.29 & 55.22 $\pm$ 1.05 & 70.19 $\pm$ 0.19 & 66.27 $\pm$ 4.08 & 49.18 $\pm$ 14.96 & 66.48 $\pm$ 9.86 & 76.54 $\pm$ 2.11 & 82.26 $\pm$ 12.33 \\
PFedCS & 78.10 $\pm$ 3.63 & \underline{92.93} $\pm$ 0.43 & \textbf{87.64} $\pm$ 0.51 & \textbf{82.38} $\pm$ 1.14 & \textbf{75.00} $\pm$ 2.00 & \underline{82.66} $\pm$ 0.89 & \textbf{82.00} $\pm$ 0.86 & \textbf{59.68} $\pm$ 1.11 & \textbf{81.43} $\pm$ 0.29 & 84.62 $\pm$ 5.80 & \underline{92.18} $\pm$ 0.93 \\
\textbf{HDC-FL} & 83.45 $\pm$ 2.03 & \textbf{95.05} $\pm$ 1.00 & \underline{86.24} $\pm$ 3.82 & \underline{81.82} $\pm$ 0.99 & \underline{74.52} $\pm$ 2.66 & \textbf{83.00} $\pm$ 0.51 & \underline{81.28} $\pm$ 1.05 & \underline{59.20} $\pm$ 0.71 & \underline{81.02} $\pm$ 0.25 & 83.75 $\pm$ 9.55 & \textbf{93.17} $\pm$ 1.61 \\
\bottomrule
\end{tabular}}
\end{table}

\begin{table}[!htbp]
\caption{Final bottom-20\% client test accuracy (\%), mean $\pm$ sample
standard deviation over three seeds.}
\label{tab:bottom}
\vskip 0.1in
\centering
\setlength{\tabcolsep}{2.5pt}
\renewcommand{\arraystretch}{1.05}
\resizebox{\linewidth}{!}{%
\begin{tabular}{@{}lccccccccccc@{}}
\toprule
Method & C10 LS2 & C10 LS3 & C10 Rot & C100 $\alpha{=}.07$ & C100 $\alpha{=}.3$ & C100 LS2 & C100 LS3 & C100 Rot & M-C100 LS3 & M-C10 $\alpha{=}.07$ & M-C10 LS3 \\
\midrule
FedAvg-SE & 92.05 $\pm$ 1.36 & 91.84 $\pm$ 0.71 & 82.45 $\pm$ 2.19 & 75.96 $\pm$ 1.24 & 68.10 $\pm$ 0.16 & 77.20 $\pm$ 0.39 & 71.29 $\pm$ 0.29 & 53.68 $\pm$ 1.77 & 73.74 $\pm$ 0.20 & 90.26 $\pm$ 1.23 & 91.91 $\pm$ 0.54 \\
FedBABU & 89.45 $\pm$ 4.26 & 91.59 $\pm$ 5.49 & 82.78 $\pm$ 1.30 & 75.71 $\pm$ 10.75 & 68.89 $\pm$ 0.82 & 64.71 $\pm$ 20.65 & 59.18 $\pm$ 18.77 & 47.08 $\pm$ 10.85 & 56.46 $\pm$ 9.34 & 88.73 $\pm$ 5.46 & 87.03 $\pm$ 7.76 \\
FedALA & 91.64 $\pm$ 2.19 & 91.58 $\pm$ 2.75 & 82.78 $\pm$ 1.97 & 74.92 $\pm$ 1.38 & 69.90 $\pm$ 1.20 & 76.92 $\pm$ 1.30 & 73.85 $\pm$ 0.01 & 59.89 $\pm$ 0.10 & 74.27 $\pm$ 1.04 & \underline{90.86} $\pm$ 0.99 & 90.67 $\pm$ 1.95 \\
FedCE & 92.11 $\pm$ 1.98 & 92.55 $\pm$ 0.37 & 81.61 $\pm$ 3.01 & 77.17 $\pm$ 0.79 & 73.39 $\pm$ 1.52 & 77.74 $\pm$ 0.62 & 74.15 $\pm$ 1.02 & 59.88 $\pm$ 0.18 & 74.99 $\pm$ 1.12 & \textbf{91.00} $\pm$ 1.45 & 92.94 $\pm$ 0.57 \\
FedSoft & 90.48 $\pm$ 1.67 & 91.20 $\pm$ 2.26 & 78.41 $\pm$ 2.87 & 77.01 $\pm$ 0.84 & 70.82 $\pm$ 1.10 & 77.36 $\pm$ 0.18 & 74.41 $\pm$ 0.36 & 56.73 $\pm$ 1.22 & 75.73 $\pm$ 0.92 & 88.32 $\pm$ 3.37 & 91.91 $\pm$ 1.09 \\
IFCA & 90.35 $\pm$ 4.31 & 86.74 $\pm$ 5.44 & 84.12 $\pm$ 1.78 & 75.84 $\pm$ 0.04 & 71.63 $\pm$ 0.09 & 74.58 $\pm$ 1.31 & 73.84 $\pm$ 2.27 & 59.78 $\pm$ 0.73 & 72.80 $\pm$ 0.63 & 73.45 $\pm$ 1.85 & 90.31 $\pm$ 0.43 \\
PACFL & 91.15 $\pm$ 1.73 & 91.39 $\pm$ 0.58 & 81.15 $\pm$ 1.00 & 77.87 $\pm$ 0.36 & 72.68 $\pm$ 0.18 & 77.52 $\pm$ 0.27 & 74.26 $\pm$ 1.16 & \underline{60.43} $\pm$ 0.93 & 75.36 $\pm$ 0.86 & 78.66 $\pm$ 9.61 & 92.75 $\pm$ 1.81 \\
EMD-CFL & \textbf{92.99} $\pm$ 0.92 & 91.26 $\pm$ 1.04 & 75.69 $\pm$ 2.82 & 73.37 $\pm$ 2.21 & 56.73 $\pm$ 1.32 & 74.55 $\pm$ 0.38 & 70.39 $\pm$ 2.41 & 51.43 $\pm$ 12.84 & 71.12 $\pm$ 5.23 & 83.07 $\pm$ 1.44 & 89.30 $\pm$ 3.65 \\
PFedCS & 88.57 $\pm$ 1.48 & \underline{95.26} $\pm$ 1.13 & \underline{88.32} $\pm$ 0.19 & \textbf{83.43} $\pm$ 0.81 & \textbf{76.78} $\pm$ 1.77 & \underline{84.87} $\pm$ 0.59 & \textbf{82.86} $\pm$ 0.18 & 60.76 $\pm$ 1.45 & \textbf{82.02} $\pm$ 0.53 & 89.54 $\pm$ 2.62 & \underline{93.83} $\pm$ 1.16 \\
\textbf{HDC-FL} & 90.51 $\pm$ 0.46 & \textbf{95.96} $\pm$ 1.09 & \textbf{88.45} $\pm$ 1.21 & \underline{83.05} $\pm$ 0.65 & \underline{76.59} $\pm$ 1.01 & \textbf{85.13} $\pm$ 0.53 & \underline{81.99} $\pm$ 0.44 & \textbf{60.99} $\pm$ 0.27 & \underline{81.69} $\pm$ 0.56 & 89.72 $\pm$ 2.40 & \textbf{94.81} $\pm$ 0.92 \\
\bottomrule
\end{tabular}}
\end{table}

\begin{table}[!htbp]
\caption{Final fairness gap (best minus worst client accuracy,
percentage points; lower is better). Each setting is first averaged
over its seeds; the last column macro-averages the 11 setting-level
means.}
\label{tab:gap}
\vskip 0.1in
\centering
\setlength{\tabcolsep}{2.5pt}
\renewcommand{\arraystretch}{1.05}
\resizebox{\linewidth}{!}{%
\begin{tabular}{@{}lcccccccccccc@{}}
\toprule
Method & C10 LS2 & C10 LS3 & C10 Rot & C100 $\alpha{=}.07$ & C100 $\alpha{=}.3$ & C100 LS2 & C100 LS3 & C100 Rot & M-C100 LS3 & M-C10 $\alpha{=}.07$ & M-C10 LS3 & Macro \\
\midrule
FedAvg-SE & 11.17 & 8.70 & 11.45 & 13.79 & 10.06 & 11.90 & 15.26 & 9.89 & 9.62 & 12.08 & 8.69 & 11.15 \\
FedBABU & 22.36 & 17.24 & 18.70 & 38.38 & 54.44 & 42.58 & 55.82 & 58.12 & 62.91 & 23.72 & 27.78 & 38.37 \\
FedALA & 12.84 & 8.05 & 11.55 & 16.47 & 11.59 & 12.13 & 9.16 & 10.51 & 9.35 & 11.13 & 8.95 & 11.07 \\
FedCE & 11.67 & 7.87 & 13.39 & 13.81 & 8.57 & 12.67 & 11.21 & 11.86 & 10.05 & 11.72 & 7.48 & 10.94 \\
FedSoft & 12.34 & 11.29 & 14.03 & 12.94 & 12.23 & 12.23 & 12.06 & 10.65 & 10.72 & 17.37 & 10.31 & 12.38 \\
IFCA & 13.54 & 28.94 & 10.52 & 13.26 & 10.80 & 14.32 & 14.13 & 12.46 & 14.31 & 52.72 & 11.91 & 17.90 \\
PACFL & 11.39 & 10.58 & 13.29 & 12.15 & 11.10 & 12.55 & 10.48 & 10.96 & 11.66 & 44.97 & 7.72 & 14.26 \\
EMD-CFL & \textbf{9.19} & 9.77 & 14.87 & 19.65 & 24.12 & 15.19 & 16.35 & 17.85 & 17.55 & 22.96 & 15.60 & 16.65 \\
PFedCS & 21.71 & 6.63 & \textbf{6.46} & \textbf{9.17} & \textbf{8.55} & \textbf{9.64} & \textbf{7.33} & \textbf{8.33} & \textbf{6.98} & \textbf{15.38} & 6.98 & 9.74 \\
\textbf{HDC-FL} & 16.55 & \textbf{4.47} & 9.14 & 10.09 & 8.80 & 9.97 & 7.80 & 10.66 & 7.16 & 16.25 & \textbf{6.08} & \textbf{9.72} \\
\bottomrule
\end{tabular}}
\end{table}

\section{ADDITIONAL TRAINING DYNAMICS AND DIAGNOSTICS}
\label{app:curves}

\begin{figure}[!htbp]
\centering
\IfFileExists{figures/curves_main.pdf}
  {\includegraphics[width=\linewidth]{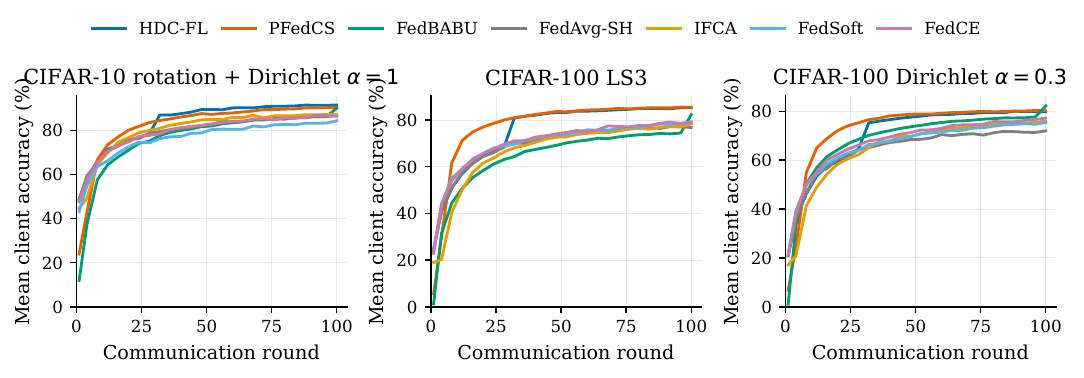}}
  {\fbox{\parbox[c][3.6cm][c]{0.98\textwidth}{\centering
   \emph{Placeholder: figures/curves\_main.pdf (three learning-curve
   panels).}}}}
\caption{Mean client test accuracy over communication rounds for
CIFAR-10 rotation, CIFAR-100 LS3 and CIFAR-100 $\alpha=0.3$, averaged
over three seeds. Before activation the HDC-FL curve reports the shared
trajectory; afterwards it reports the personalized prototype mixtures
constructed from it.}
\label{fig:curves}
\end{figure}

\paragraph{Training dynamics.} Figure~\ref{fig:curves} shows mean
client accuracy over communication rounds for three representative
settings. Before activation, HDC-FL follows the shared trajectory;
afterwards, the curve reports the personalized prototype mixture. The
activation step is most visible on CIFAR-10 rotation and CIFAR-100 LS3
and smaller on CIFAR-100 $\alpha=0.3$. After activation, performance
remains stable because prototypes are rebuilt from the current cache at
each refresh. These curves describe round-wise predictive accuracy
rather than optimization speed, since activation changes the evaluated
model but not the underlying training procedure.

\begin{figure}[!htbp]
\centering
\IfFileExists{figures/learning_curves_FedAvgsharedencoder.pdf}
  {\includegraphics[width=0.97\linewidth]{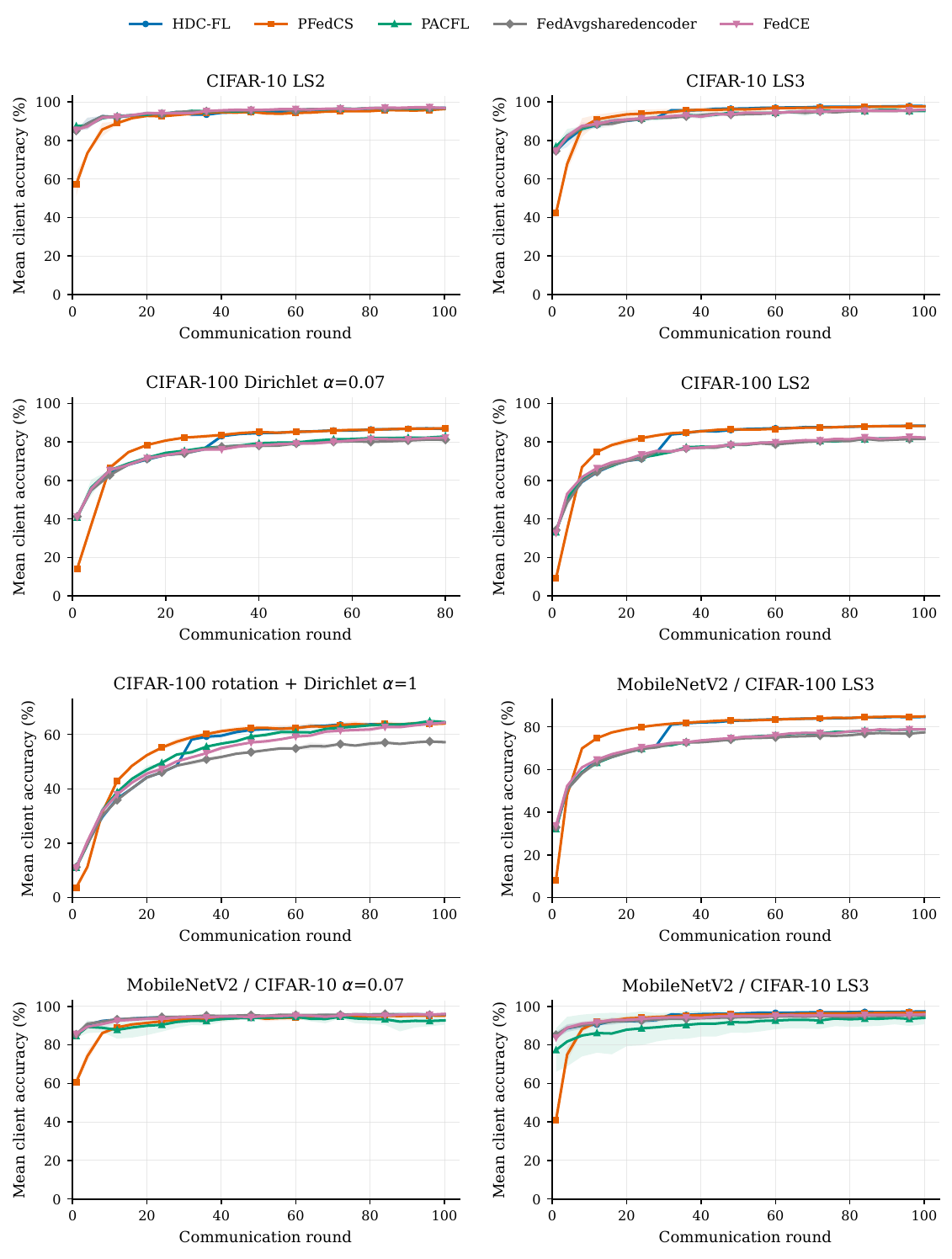}}
  {\fbox{\parbox[c][4cm][c]{0.98\linewidth}{\centering
   \emph{Placeholder: figures/curves\_extra.pdf (learning curves for the
   remaining configurations).}}}}
\caption{Learning curves for the configurations not shown in
Figure~\ref{fig:curves}, averaged over three seeds. Shaded regions are
one sample standard deviation.}
\label{fig:curves-extra}
\end{figure}

Figure~\ref{fig:curves-extra} extends Figure~\ref{fig:curves} to the
remaining configurations. In every case HDC-FL follows the shared
trajectory until activation and then improves or remains level, and the
size of the step at activation is larger in the label-subset and
rotation settings than in the Dirichlet ones.

\begin{figure}[!htbp]
\centering
\IfFileExists{figures/diagnostics.pdf}
  {\includegraphics[width=\linewidth]{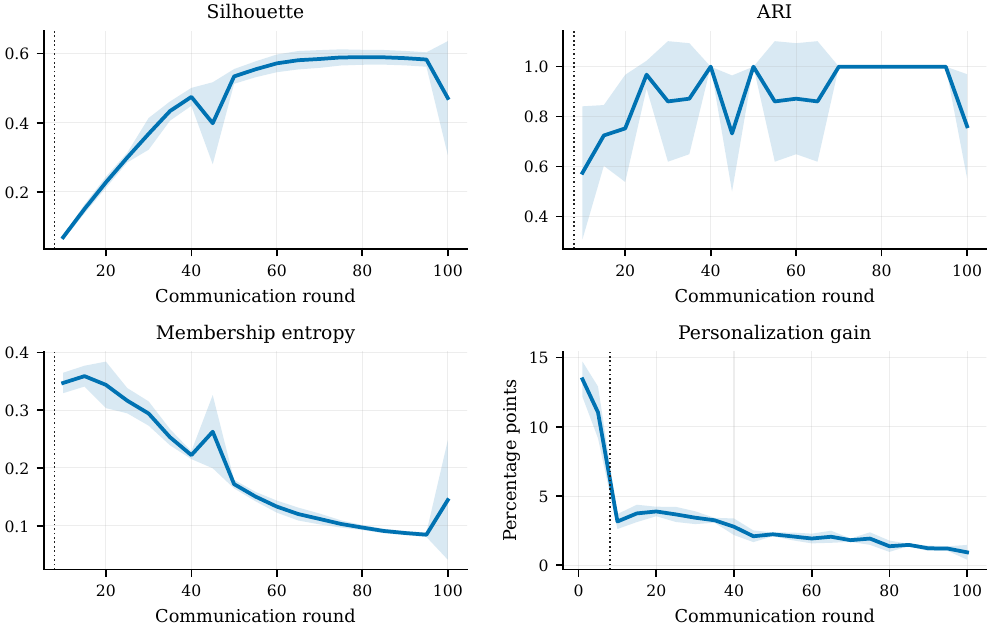}}
  {\fbox{\parbox[c][4cm][c]{0.98\linewidth}{\centering
   \emph{Placeholder: figures/diagnostics.pdf (silhouette, ARI,
   membership entropy, personalization gain).}}}}
\caption{Grouping diagnostics over communication rounds on CIFAR-10
rotation without pretrained initialization with four groups: cluster
separation (silhouette), agreement between consecutive assignments
(ARI), membership confidence (entropy), and personalization gain.
Lines and shaded regions are the mean and one standard deviation over
three seeds.}
\label{fig:diag}
\end{figure}

Figure~\ref{fig:diag} reports the internal diagnostics of the grouping
over training on CIFAR-10 rotation without pretrained initialization.
The silhouette score increases as the descriptor space becomes more
structured, the adjusted Rand index between consecutive refreshes stays
high once the grouping has settled, and the membership entropy falls,
indicating that clients become more confidently associated with one
prototype. The personalization gain is largest immediately after
activation and decreases as the shared encoder catches up, which is
consistent with the flattening of the curves in
Figure~\ref{fig:curves}.

\clearpage
\section{ADDITIONAL ABLATIONS AND DESCRIPTOR DIAGNOSTICS}
\label{app:sens}

\subsection{Low-Rank Descriptor}
\label{app:desc}

Table~\ref{tab:abl-desc} compares the rank-8
descriptor with the raw classifier head on CIFAR-100 LS3, where the
class dimension is large enough for the projection to act. Accuracy is
unchanged (85.24\% against 85.27\%), so the projection is not what
produces the predictive performance. What it changes is the geometry of
the descriptor space: the silhouette score rises from 0.027 to 0.108
and the agreement between the assignments of consecutive refreshes,
measured by the adjusted Rand index, rises from 0.152 to 0.401. Groups
found on raw heads are therefore poorly separated and unstable over
time, while the projected descriptors give a grouping that persists
between refreshes; this is an argument for stability, not for
accuracy.

\begin{table}[!htbp]
\caption{Descriptor. Sil.\ is the silhouette score of the resulting
clustering and ARI the agreement between the assignments of consecutive
refreshes; both are higher-is-better. Means $\pm$ sample standard
deviations over three seeds.}
\label{tab:abl-desc}
\vskip 0.1in
\centering

\resizebox{\columnwidth}{!}{%
\begin{tabular}{@{}lccc@{}}
\toprule
Descriptor & Mean & Sil.\  & ARI  \\
\midrule
Raw head
& \textbf{85.27} $\pm$ 0.09
& 0.027 $\pm$ 0.002
& 0.152 $\pm$ 0.065 \\

Low rank ($r{=}8$)
& 85.24 $\pm$ 0.31
& \textbf{0.108} $\pm$ 0.012
& \textbf{0.401} $\pm$ 0.080 \\
\bottomrule
\end{tabular}%
}
\end{table}

\subsection{Descriptor Rank}
\label{app:rank}

Table~\ref{tab:rank} varies the descriptor rank on CIFAR-100 rotation
without pretrained initialization, using a single seed. Predictive
performance is essentially flat over $r\in\{8,20,40,60,80\}$, varying
by less than one point, so the rank is not an accuracy-critical
hyperparameter. The silhouette score, however, decreases monotonically
with $r$: retaining more class-space directions produces a descriptor
space whose clusters are less separated. Predictive performance changes by less than one percentage point across the ranks examined, whereas the silhouette score decreases as the retained rank increases. Among the values evaluated in this diagnostic, the fixed \(r=8\) configuration used in the main experiments yields the highest silhouette score.

\begin{table}[!htbp]
\caption{Descriptor-rank sensitivity, single seed.}
\label{tab:rank}
\vskip 0.1in
\centering
\small
\begin{tabular}{@{}ccccc@{}}
\toprule
Rank $r$ & Mean & Worst & Bot.20\% & Silhouette $\uparrow$ \\
\midrule
8  & 40.85 & \textbf{35.57} & \textbf{36.71} & \textbf{0.195} \\
20 & \textbf{40.96} & 35.38 & 36.36 & 0.126 \\
40 & 40.24 & 34.02 & 35.91 & 0.090 \\
60 & 40.50 & 34.05 & 35.71 & 0.068 \\
80 & 40.23 & 34.39 & 35.84 & 0.059 \\
\bottomrule
\end{tabular}
\end{table}

\subsection{Prototype Membership}
\label{app:mem}

Table~\ref{tab:abl-mem} replaces the soft
mixture by a hard assignment to the most similar prototype, on
CIFAR-100 with $\alpha=0.3$. Soft mixing is better on all three
metrics, by 0.85 points on the mean and 1.08 and 0.95 points on the
two tail metrics. This is the setting in which client distributions
form a continuum rather than discrete groups, and the tail improves
more than the mean---what one expects if the clients that benefit are
those whose descriptors lie between groups and are served badly by any
single one.

\begin{table}[!htbp]
\caption{Prototype membership. Means $\pm$ sample standard deviations
over three seeds.}
\label{tab:abl-mem}
\vskip 0.1in
\centering
\footnotesize
\setlength{\tabcolsep}{3pt}
\begin{tabular}{@{}lccc@{}}
\toprule
Membership & Mean & Worst & Bot.20\% \\
\midrule
Hard assignment & 78.89 $\pm$ 0.46 & 73.55 $\pm$ 1.18 & 75.39 $\pm$ 0.52 \\
Soft mixture & \textbf{79.74} $\pm$ 0.40 & \textbf{74.63} $\pm$ 0.26 & \textbf{76.34} $\pm$ 0.46 \\
\bottomrule
\end{tabular}
\end{table}

\subsection{Mixing Coefficient}
\label{app:lambda}

We vary $\lambda$ on CIFAR-100 LS3 as a single-seed diagnostic. At
$\lambda=0$ the client retains its cached encoder and obtains 76.95\%
mean accuracy, consistent with FedAvg-SE in the same setting.
Performance increases to 83.72\% at $\lambda=0.5$ and 85.22\% at
$\lambda=1$, with worst-client and bottom-20\% accuracy following the
same ordering. Within this diagnostic, \(\lambda=1\) gives the highest observed mean, worst-client, and bottom-20\% accuracy among the three values examined; the main experiments use the same fixed \(\lambda=1\) configuration throughout.

\begin{table}[!htbp]
\caption{Mixing coefficient $\lambda$. Single-seed diagnostic.}
\label{tab:abl-lambda}
\vskip 0.1in
\centering
\footnotesize
\setlength{\tabcolsep}{3pt}
\begin{tabular}{@{}cccc@{}}
\toprule
$\lambda$ & Mean & Worst & Bot.20\% \\
\midrule
0.0 & 76.95 & 71.93 & 72.58 \\
0.5 & 83.72 & 79.92 & 80.34 \\
1.0 & \textbf{85.22} & \textbf{80.12} & \textbf{81.77} \\
\bottomrule
\end{tabular}
\end{table}

\section{DONOR-COMPATIBILITY DIAGNOSTIC}
\label{app:donor}

This appendix gives the full protocol and the additional measurements
behind Section~\ref{sec:donor}.

\paragraph{Protocol.} The diagnostic is run at the final round
($t=100$) of MobileNetV2 training with 20 clients, full participation
and seeds 42, 123 and 2026, under three settings: CIFAR-10 label-subset
skew, CIFAR-100 label-subset skew (both with 30\% label support per
client) and CIFAR-100 rotation with $\mathrm{Dir}(1)$ label
proportions, where clients 0--4, 5--9, 10--14 and 15--19 receive
$0^{\circ}$, $90^{\circ}$, $180^{\circ}$ and $270^{\circ}$
respectively. For each ordered pair $(t,d)$ with $d\neq t$, the target
is first evaluated with its own encoder and head to obtain
$\mathrm{self\_loss}(t)$ and $\mathrm{self\_acc}(t)$; the target
encoder is then replaced by donor $d$'s cached encoder, the target head
is left unchanged, and the hybrid model is evaluated on the same target
test split to obtain $\mathrm{donor\_loss}(t,d)$ and
$\mathrm{donor\_acc}(t,d)$. The loss benefit is
$\mathrm{self\_loss}-\mathrm{donor\_loss}$ and the accuracy gain is
$\mathrm{donor\_acc}-\mathrm{self\_acc}$, both positive when the
transfer helps. Nothing is retrained or fine-tuned after the swap.
Pairs are directed, since target data and target head differ between
the two orderings, giving $20\times 19=380$ pairs per run and
$3{,}420$ in total. All correlation and selection metrics are computed
per target over its 19 candidate donors, averaged over the 20 targets
within a seed, and then averaged over seeds; sample standard deviations
are taken across the three seeds. Distance-quintile summaries instead
treat the 60 target$\times$seed units of a setting as the analysis
units and report standard errors. The final mean client accuracies of
these runs (97.45\%, 84.67\% and 61.55\%) match the corresponding
MobileNetV2 entries of Table~\ref{tab:mobilenet} for the two settings
that also appear there.

\paragraph{Distance quintiles.} For each target, donors are ordered by
low-rank head distance and split into five groups, Q1 being the closest.
Table~\ref{tab:donor-quintiles} reports the accuracy gain per quintile.
The ordering is monotone or nearly monotone in all three settings,
which supports the ranking interpretation without assuming a linear
relationship. The absolute values also make the scale of the
intervention clear: on CIFAR-10 label skew and CIFAR-100 rotation a
single encoder swap costs accuracy on average in every quintile, and
head proximity reduces that cost rather than turning it into a gain.

\begin{table}[t]
\caption{Accuracy gain (percentage points) of a single donor-encoder
swap, by low-rank head-distance quintile; Q1 is the closest fifth of
donors. Values are means over the 60 target$\times$seed units of a
setting, with standard errors.}
\label{tab:donor-quintiles}
\vskip 0.1in
\centering
\small
\begin{tabular}{@{}lccccc@{}}
\toprule
Setting & Q1 & Q2 & Q3 & Q4 & Q5 \\
\midrule
C10 LS3   & $-2.96$ & $-3.65$ & $-4.11$ & $-4.51$ & $-4.63$ \\
         & \scriptsize(0.35) & \scriptsize(0.41) & \scriptsize(0.42) & \scriptsize(0.42) & \scriptsize(0.55) \\
C100 LS3  & $+0.99$ & $+0.97$ & $+0.86$ & $+0.55$ & $+0.47$ \\
         & \scriptsize(0.27) & \scriptsize(0.26) & \scriptsize(0.25) & \scriptsize(0.24) & \scriptsize(0.26) \\
C100 Rot & $-1.16$ & $-2.69$ & $-2.99$ & $-2.95$ & $-3.13$ \\
         & \scriptsize(0.32) & \scriptsize(0.29) & \scriptsize(0.23) & \scriptsize(0.26) & \scriptsize(0.26) \\
\bottomrule
\end{tabular}
\end{table}

\paragraph{Rotation structure.} In the rotation setting the true client
groups are known, so head similarity can be evaluated directly as a
group detector. The nearest-head donor shares the target's rotation in
$75.0\%$ of cases against a base rate of $21.1\%$; head similarity
detects same-rotation pairs with AUROC $0.887$ and average precision
$0.606$; and mean head similarity is $0.789$ within a rotation group
against $0.752$ across groups. Transfer follows the same split:
same-rotation swaps give a mean loss benefit of $+0.072$ and $+0.83$
accuracy points, different-rotation swaps $-0.162$ and $-3.46$ points.

\paragraph{Label-support overlap.} In the two label-skew settings we
compare head similarity with the Jaccard overlap of the clients'
observed label supports. The two are strongly associated (Spearman
$0.888\pm0.058$ on CIFAR-10 and $0.779\pm0.097$ on CIFAR-100). After
residualizing the overlap ranks, the partial Spearman correlation
between head similarity and loss benefit is $-0.092\pm0.265$ on
CIFAR-10 and $0.090\pm0.270$ on CIFAR-100, i.e.\ indistinguishable from
zero at this sample size. We therefore read head similarity in the
label-skew regime as largely reflecting shared label support. This
partial correlation is exploratory and is not a causal estimate.

\paragraph{Low-rank descriptor versus raw head.} All target-wise
metrics were recomputed with raw head similarity in place of the
rank-8 descriptor (Table~\ref{tab:donor-desc}). The low-rank descriptor
is clearly better only on CIFAR-100 label skew, and is marginally worse
in the other two settings, so these results do not support a claim that
the low-rank projection is universally more predictive of transfer.
This is consistent with Section~\ref{sec:ablation} and
Appendix~\ref{app:sens}, where the projection improves the separation
and temporal stability of the grouping rather than predictive accuracy.

\begin{table}[t]
\caption{Low-rank versus raw head similarity as a donor-ranking signal.
Positive differences favour the low-rank descriptor.}
\label{tab:donor-desc}
\vskip 0.1in
\centering
\small
\begin{tabular}{@{}lccc@{}}
\toprule
 & C10 LS3 & C100 LS3 & C100 Rot \\
\midrule
Spearman $\rho$, low rank & 0.193 & 0.117 & 0.243 \\
Spearman $\rho$, raw      & 0.199 & 0.050 & 0.288 \\
Difference                & $-0.006$ & $+0.067$ & $-0.045$ \\
\midrule
AUROC, low rank           & 0.665 & 0.565 & 0.783 \\
AUROC, raw                & 0.677 & 0.502 & 0.793 \\
Difference                & $-0.012$ & $+0.064$ & $-0.010$ \\
\bottomrule
\end{tabular}
\end{table}

\paragraph{Scope.} The diagnostic measures zero-shot encoder
interchangeability conditional on a fixed target head at one point in
training, on one backbone. It establishes an association and a
donor-ranking result, not a causal or general claim about encoder
equivalence, and it does not evaluate HDC-FL's prototypes themselves,
which average several cached encoders rather than substituting one.

\end{document}